# Collocational Grammar

R. J. Freeman
Language Centre
Hong Kong University of Science and Technology
lcrobf@usthk.ust.hk

**Abstract**

A perspective of statistical language models which emphasizes their collocational aspect is advocated. It is suggested that strings be generalized in terms of classes of relationships instead of classes of objects. The single most important characteristic of such a model is a mechanism for comparing patterns. When patterns are fully generalized a natural definition of syntactic class emerges as a subset of relational class. These collocational syntactic classes should be an unambiguous partition of traditional syntactic classes.

## Introduction

I think one can characterize the fundamental model of language structure as having always been that of classes of objects related in more or less simple ways. I propose that it might be useful to think instead in terms of classes of relationships of more or less simple objects. This corresponds to giving priority to collocation (e.g. Lewis 1993) rather than syntax as the central mechanism underlying language structure.

An assertion of primacy for relationships rather than objects may not seem so world shaking. Any model of structure has both aspects. Nevertheless, I think that an "object based" perspective has dominated in NLP to the detriment of our understanding of what it is about our models which make them work, and thus how they can be improved. Giving primacy to relationships need not so much change what we do but more importantly cause a re-evaluation of why, (when, and how much) we do it. This could lead to the resolution of some central problems.

## Relation based language models

Let us look at some current language modelling techniques from a relational perspective.

### N-gram models

N-gram models, which look at word groupings rather than lexical classes can be considered to be relational models of language. They have indeed proven the most useful in recent times, though I doubt whether many of their proponents would agree that lack of lexical classification is the reason for their success. Certainly their success cannot be attributed to generalization of relationships, because none takes place, however, there is a change of focus when compared to earlier approaches to modelling language structure, in n-gram models relationships play an equal role in terms of representational complexity. In fact, this almost happens by default. N-gram models have not so much been motivated by the positive desire to model relationships, but by the negative need to deal with runaway ambiguity in lexical classes (Church, 1988).

Whatever the motivation, however, the fact remains, in n-gram models relationships are given an importance comparable to lexis for the first time.

A problem is the simple format of n-gram models requires that an explicit record of all relationships be kept. The central processing problem becomes one of data storage and collection. It is no accident that n-gram models are closely associated with electronic corpora, and increases in electronic data storage efficiency.

### Data storage

It is interesting to look at the efficiency of an n-gram model in terms of a data storage problem. I argue that for a relational model practical issues of data storage, and theoretical issues of representational accuracy become one. Perhaps this is so of any model. After all the essential property of any structural description is that it should capture a maximum of structure in a minimum of description.

**Locality of information.** In the case of n-grams there is no generalization between relationships. So to minimize the data storage problem examples are kept to a reasonable length. If the length is restricted many sub-strings will be repeated. These repeated strings need only be stored once, and data bases can be kept to a reasonable size.

The storage space needed to store all the patterns increases exponentially for each additional token, so if long range information was important impossibly large data bases would still be required to create and store the n-grams needed to describe a reasonable variety of language. Fortunately, when comparisons are made between the information modelled by n-grams and that in naturally occurring language, it is found that even if example strings are as short as 2 or 3 tokens most information is captured. So most of the information is quite local, and modelling language in this way becomes a practical proposition.

The problem is that while the information contained in longer range dependencies becomes vanishingly small in general, in particular cases it is crucial, and if no generalization takes place this information is impossible to collect or store.

**Decision-Trees**

Data storage in the n-gram model is inefficient because it stores everything about each unique pattern met. Normally, however, there will be similarities between patterns, it would be nice if we could take advantage of that.

A technique which has been used recently in this way is the Decision-Tree (Magerman 1994). A Decision-Tree stores all historical similarities in a set of data. That is all common structure between the patterns, *up to a given branch point*, is stored only once. In this way many different patterns can be stored with only modest increases in storage space. This is extremely useful when a large number of observed patterns share a common history[1]. An example of the relative economies of the two systems is shown below:

Information to be stored as n-grams

A B C D E F G H I J K L M
A B C D E F G I H J K L M

Information to be stored using a Decision-Tree

A B C D E F G H I J K L M
　　　　　　　＼
　　　　　　I H J K L M

In Magerman's work the motivation for the use of a tree structure is storage efficiency. As a side effect he also classifies an infinity of patterns. For instance, the infinity of patterns which begin with A, AH, BFG, or any other combination of tokens.

The power this infinity of patterns bestows on him is responsible for the success of his model. By classifying it he has generalized a large amount of information about important relationships in his data, and it is on the information in these relationships that the accuracy of his parsing decisions is based, just as it is in an n-gram model. His objective is not to classify relationships, however, he still sees parsing as "making a sequence of disambiguation decisions" about the objects in his structure, not as making a disambiguation decision about his structure as a whole. The essential abstraction of information in his analysis is supposed to take place within each branch of the tree, not in terms of a comparison between them.

---

1. It has been argued that languages can be divided into post-modifying (e.g. French) and pre-modifying (e.g. German). (English is supposed to be in the process of changing from pre to post modifying.) It might be interesting to compare the usefulness of Decision-Trees for storing patterns in pre and post modifying languages, as history would give the modification in one case but not in the other.

As in the case of the n-gram models relationships are conceived of in the Decision-Tree formalism only as an aid disambiguation of syntactic categories. Once again we see more emphasis on relationships coming about as a side effect of a proliferation of problems with object classes. Yet it is these object classes which are supposed to be simplifying the structure.

**Factor model**

Magerman is content to generalize only historical similarities in the relationships between tokens in his structures because he assumes that the essential generalization of information will occur within his token classes. Despite this emphasis on syntactic classes, however, a pattern of an increase in storage efficiency associated with more complete classification of patterns, is clear from the example of the Decision-Tree. We can see that even greater efficiencies would be possible if we could store not only the historical similarities, but *all* the common sub-structure between patterns found in the data.

An example of a technique which attempts to extract all the similarities in a distribution of data, a random field, is the "incremental feature induction algorithm" of Della Pietra et al (1995). Their technique is presented in the form of a general representation of regularities in any "random" string of objects. The particular example they use is English orthography.

In their analysis, effectively, a string of tokens is segmented into all common sub-strings, and it is these which are stored the one time necessary for complete representation of the data. The sub-strings combine like irreducible multiplicative factors to form larger strings (this multiplicative combination is explicit when the probabilities of the respective sub-strings are combined to estimate the probability of a complete string), thus this analysis can be thought of as a reduction to a "factorial code" (Redlich, 1993) of the set of all possible strings.

The situation for our example of two 13 token strings is given below:

Information stored using Decision-Tree

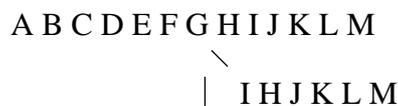

Information stored using the feature model of Della Pietra et al.

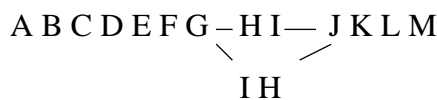

By storing similarities over the whole string such a model manages to represent (almost, see "Hidden structure") all the information in the original two 13 token strings, in four strings, containing a total of only 15 tokens. For representing the relationships between successive members of a string of objects then, the factor model is better than the n-gram model, where common sub-strings are not identified at all, and the Decision-Tree model, where only historical sub-strings are identified.

**Storage efficiency and classification efficiency**

For a given length of pattern we can see how similarity measures can be used to increase storage efficiency as below:

- - - ➤ storage required by n-gram

———➤ storage required by fac. model

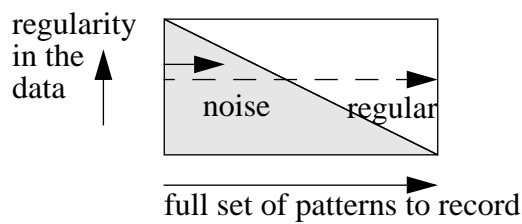

In general, a set of data is divided into regular and noisy forms. Moving from n-grams to a full factor model the storage emphasis moves from storing all of the data, to storing only the differences in the data. If the data is highly regular then even for long patterns it becomes practical to record all the forms (If the data is random then the storage requirements of the factor model are the same as those of the n-gram model).

The model which gives the best classification of similarity in the relationships between successive tokens will be the one which stores the regularities in the strings most effectively, i.e. a factorial reduction along the lines of Della Pietra et al.

## Grammatical Representation

So we see that there is a second perspective for making grammatical generalizations. That of generalizing the relationships rather than the objects. We can see that the use of relationships has been implicit in our most effective practical modelling tools, though the information has been used almost reluctantly, as a last resort, forced on us by vagaries of the token classes. Nevertheless, the influence of relationships has increased, and as it has increased researchers have looked for ways to store more of the information they provide. That has meant classifications of relationships have crept into their analyses.

The sequence of analyses listed above demonstrate that the most efficient way to store the information in any distributed data, and relationships between many tokens are by their nature distributed, is in terms of factors. This is what is done in the analysis of Della Pietra et al. for orthography. But there is a discontinuity in my sequence of examples. The n-gram and Decision-Tree examples are specifically applied to modelling syntax, but not the factorial reduction. Della Pietra et al. recognize that a full factorial reduction of their data is needed for strings of letters, why has this not been applied to the modelling of patterns in syntax by other researchers?

### Time

I think it is the influence of the "object based" perspective which has limited us again. It is not recognized that the pattern of relationships in the string as a whole should be taken into account in linguistic decisions. Magerman, for instance, is still thinking in terms of sequences of decisions about token classes. Why should something which has not yet happened influence a classification which is occurring now. It is reasonable from the perspective of a series of decisions which take place in a strict chronological order that only historical events should be allowed to contribute. He assumes that historical generalizations will be enough. When one explicitly identifies relationships rather than tokens as the central defining characteristic of structure, however, we no longer need think about any particular token, at a particular time in the sequence, and it is less obvious that time should be a factor. In a relationship based model, all the tokens in a string contribute to a classification, and it becomes clear that we need to extract all the generalities in the string to make a complete generalization, not just the historical ones.

### Syntactic category

The change in perspective associated with recognizing relationships as the basis of classification leads us, literally, to whole new classes of information. What then of our familiar lexical classes, are they to be abandoned entirely? In fact relationship classes give us a definition for lexical generality which explains the identity, and also the ambiguity of our traditional definitions.

**Hidden Structure.** Della Pietra et al. talk of hidden structure when they discuss extensions to their factor analysis of a random string. By this they mean variation within the basic framework of a given sub-string. Take a group of letters like "bed" in English, for example. In the analysis of Della Pietra et al. this should be isolated as an irreducable information bearing element. However, there will be other elements similar to it, "bud", "bid", "bad", etc. The information in the common structure of all these words is hidden inside them. In the case of strings of words this hidden structure would generalize to just the syntactic classes we traditionally seek.

Seen in the light of a relational model, then, syntactic classes are best regarded as a subset of relational categories, the category of relationships between elements of a substring. The difference between these categories and those we normally use is that they will be defined only in the context of the various sub-strings with which they are associated, not independently of them.

In effect, we are led by the relational model to define a separate set of syntactic categories for each distinct collocation in our data. This may seem overly laborious, but need not be more so than the parallel task of finding a distinct set of relationships for each syntactic category as in "object based" grammars.

**Traditional classes and ambiguity.** This hidden structure is of the same form as our traditional lexical classes. It should be possible to enlarge this to give an arbitrarily exact match to the traditional classes simply by reducing the specification of the relational classes. The "identity" of traditional lexical category, will correspond to the similarity, at whatever level, of the relational classes in which the words of the category participate. Exact similarity corresponds to exactly the full relational classes, less similar matches will give progressively more general characterizations of syntactic behaviour.

For example, if the lexical class associated with a collocation like "____ wood blocks under the car." is "Place", "Put", Position", "Locate", etc. and that is combined with those associated with similar collocations like "____ the car through the tunnel." then the group of words specified will grow to include more "verbs". A class defined on the similarities of all such expressions would be one section of a general "verb" class. The more general the similarities the more general the class.

A definition in terms of relational classes also gives us a natural explanation for ambiguity, and a means for its resolution. Having similarities does not preclude having differences. Just because a word can have a "verb" like character with one set of collocations, that does not mean it cannot have a "noun" like character with another set. We can define a distinction between classes based on the differences between our collocational environments just as we can define an identity based on the similarity.

For the case of our example above, just as "Place" can be a "verb" because it is found in a verb like context, similarly it can be a "noun" because it is found in "noun" like contexts like "The ____ I live." In fact it will be a "noun" or a "verb" exactly when it is found in these respective contexts.

Not only does a definition of syntactic class in terms of relational classes solve the problem of ambiguity, it also motivates a cause. It is because distinction is based on context that the token itself can be ambiguous with respect to class. It is ambiguous because there is no reason for it not to be.

**An example of a lexical relational class.** An example of the interplay between contextual similarities and differences can be seen in the derivation of syntactic category from first principles by Schuetze (1993). Schuetze approaches the problem of classifying linguistic behaviour explicitly as one of generalizing relationships between tokens. The problem is that his context on which his classification is based is restricted to relationships at two or three tokens distance, so it does not distinguish differences beyond that range. His representations are relational, but the balance shifts from enumerating the differences between relationships to enumerating their similarities. He loses many features distinguishing the behaviour of different words, so the classes defined by his distributions are more general, but this also means the distributions are less able to resolve many ambiguities of classification. From the perspective of the relational model we know that, in general, this lost information may be at the lexical level only, and impossible to recover in terms of general classes.

Schuetze's analysis is interesting, however. It can be thought of as approaching a rela-tional model from the other direction. His focus is still on lexical classes, but he seeks to increasingly restrict his lexical classes in terms of their contexts. In the limit he would end up specifying (severely restricted) classes in terms of the collocational factors we derive directly in a general relational model.

**Higher structure.** The analysis of hidden structure would not be limited to generalizations about single tokens. Groups of tokens could have common behaviour in a wider context. These would provide a natural definition of syntactic categories above the lexical level, noun phrases, relative clauses etc. Once again an important difference is that any generalizations of tokens could be defined only in terms of a given sub-string. A generalization of the properties of these sub-strings should give us exactly our traditional syntactic categories, but the information in each of them would be greater than that of the combined category. The extra information should be exactly what we need to interpret the use of a given grouping of tokens in a particular context.

**Full generalization**

In terms of our earlier simple example of data storage efficiency the ideal situation, including a classification of hidden structure, would be something like the diagram below:.

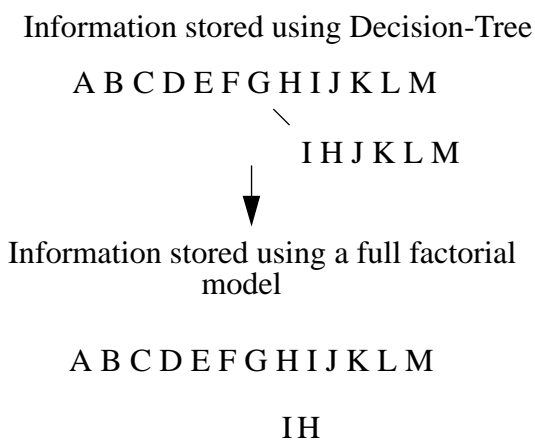

Information stored using Decision-Tree

A B C D E F G H I J K L M

I H J K L M

Information stored using a full factorial model

A B C D E F G H I J K L M

I H

The original strings, stored explicitly as two strings of 13 characters in the n-gram model, and stored as strings of 13 and 6 tokens in the Decision-Tree model, now find complete representation as two factors, one with 11 characters and the other with just two components of two characters each. I have deliberately made the elements of the hidden structure two tokens long to indicate that segmentation has taken place.

## Summary of relational models

We can summarize the important features of a relation based language model as those that enable us to store the most information about allowable sequences of tokens in our language, and to relate this information to general strings most completely. In brief:
- From a practical standpoint - storage efficiency
- From a theoretical standpoint - flexible generalizability

In fact both come down to having a means of comparison. The most effective means of comparison gives the most effective generalizability and thus the most efficient data storage. The factor model gives you the most effective means of comparison because it extracts all the similarities between the strings.

Before we can harness this power however we have to recognize that the representational complexity of the whole string must be bought to bear on each decision. This is clear when we recognize the primacy of relationships. In terms of relationships we give preference to no particular token with which to associate a particular time, and the distinction between "history" and context becomes irrelevant.

Within that context collocation comes to the fore as the fundamental mechanism of grammatical classification. Syntactic classes have a natural definition as special cases of collocational classes. Our traditional syntactic classes can be seen as a generalization of the common properties of groups of these "collocational" syntactic classes.

## Conclusion

Sometimes it seems researchers have been moving forwards backwards. It is as if everyone is fighting a fierce rearguard action struggling towards the objective of a discrete syntactic classes, all the while being forced by practical necessity to move closer and closer toward an analysis of relationships whenever they want to get anything done. Everyone is still trying to find relationships of classes rather than classes of relationships.

In n-gram models relationships are used, but not classified, and enormous data requirements make their direct interpretation impractical. In Magerman's history based parsing he effects the classification of one half of the infinity of relationships he seeks to use, but the use of relationship classes is implicit rather than explicit. Schuetze derives relationship classes, but only within a lexical not a syntactic context[1].

Generalizations are supposed to simplify problems, not create them. The lexical classes of traditional analyses are cognitive classes. They may properly be the objectives of our analyses, they need not be the means. In practice they seem to cause more problems than they solve. It is the thesis of this paper that this is because they are dependent on a more fundamental classification, that of relationships or collocational structure. I have sought to show that it is relationship classes which underlie the success of "data based" language models such as n-gram and statistical parsers, and that the most efficient way of modelling relationship classes is in terms of an analysis which factors out the greatest number of similarities in different token strings.

Rather than trying to extrapolate from lexical generality to structural generality I feel we should be moving from structural generality to lexical and syntactic generality. We can still have our familiar syntax categories but only in the context of a sub-class of the wider collocational classification. The central issue of NLP becomes, not the efficient classification of parts of speech, but of collocational regularity, the single most important tool in the analysis of language structure, an effective means of modelling similarities in strings.

---

1. Others are similarly led to "relationship classes" as a means of resolving lexical ambiguity, e.g. Yarowsky (1993). The "supertags" of Joshi and Srinivas (1994) seem very close to relationship classes though their formulation is still strongly influenced by concepts of lexical generalization.

## References

Church, K. W., A Stochastic Parts Program and Noun Phrase Parser for Unrestricted Text, ICASSP-89: 1989 International Conference on Acoustics, Speech and Signal Processing, vol. 4 p.2833.

Della Pietra, S., Della Pietra, V., and Lafferty, J., Inducing features of random fields, cmp-lg/9506014.

Joshi, A. K., Srinivas, B., Disambiguation of Super Parts of Speech (or Supertags): Almost Parsing, Proceedings of the 15th International Conference on Computational Linguistics (COLING '94), Kyoto, Japan, August, 1994.

Lewis, M., The Lexical Approach The State of ELT and a Way Forward, Language Teaching Publications, 1993.

Magerman, D. M., Statistical Decision-Tree Models for Parsing, cmp-lg/9404030.

Redlich, A. N., Redundancy Reduction as a Strategy for Unsupervised Learning, Neural Computation 5, 289-304 (1993).

Schuetze, H., Distributional Part-of-Speech Tagging, URL: ftp://csli.stanford.edu/pub/prosit/DisPosTag.ps.

Schuetze, H., Distributed Syntactic Representations with an Application to Part-of-Speech Tagging, 1993 IEEE International Conference on Neural Networks, p1504-9 vol. 3.

Yarowsky, D., One Sense Per Collocation, In Proceedings, ARPA Human Language Technology Workshop, Princeton, pp. 266-271, 1993.